\documentclass[journal]{IEEEtran}
\usepackage{cite}
\usepackage{hyperref}

\usepackage{graphicx}
\ifCLASSINFOpdf
\else
\fi
\usepackage{amsmath}
\usepackage{array}

\usepackage{gensymb}
\usepackage{booktabs}
\usepackage{subcaption}

\begin{document}
%
\title{UAV LiDAR Point Cloud Segmentation of A Stack Interchange with Deep Neural Networks}
%
%
%

\author{Weikai Tan,~\IEEEmembership{Student Member,~IEEE,}
        Dedong Zhang,
        Lingfei Ma,
        Ying Li,
        Lanying Wang,
        and~Jonathan~Li,~\IEEEmembership{Senior Member,~IEEE}
\thanks{Manuscript received xxx, 2020; revised xxx. (Corresponding author: Jonathan Li)}
\thanks{W. Tan, Y. Li and L. Wang are with the Department of Geography and Environmental Management, University of Waterloo, Waterloo, ON N2L 3G1, Canada (email: \href{mailto:weikai.tan@uwaterloo.ca}{weikai.tan@uwaterloo.ca}; 
\href{mailto:y2424li@uwaterloo.ca}{y2424li@uwaterloo.ca};
\href{mailto:lanying.wang@uwaterloo.ca}{lanying.wang@uwaterloo.ca})}
\thanks{D. Zhang is with the Department of Systems Design Engineering, University of Waterloo, Waterloo, ON N2L 3G1, Canada (email: \href{mailto:dedong.zhang@uwaterloo.ca}{dedong.zhang@uwaterloo.ca})}
\thanks{L. Ma is with the Engineering Research Center of State Financial Security, Ministry of Education and School of Information, Central University of Finance and Economics, Beijing, 100081, China (email: \href{mailto:mlfcugb@gmail.com}{mlfcugb@gmail.com})}%
\thanks{J. Li is with the Department of Geography and Environmental Management and the Department of Systems Design Engineering, University of Waterloo, Waterloo, ON N2L 3G1, Canada (email: \href{mailto:junli@uwaterloo.ca}{junli@uwaterloo.ca})}
}

\maketitle

\begin{abstract}
Stack interchanges are essential components of transportation systems. 
Mobile laser scanning (MLS) systems have been widely used in road infrastructure mapping, but accurate mapping of complicated multi-layer stack interchanges are still challenging. 
This study examined the point clouds collected by a new Unmanned Aerial Vehicle (UAV) Light Detection and Ranging (LiDAR) system to perform the semantic segmentation task of a stack interchange. 
An end-to-end supervised 3D deep learning framework was proposed to classify the point clouds. 
The proposed method has proven to capture 3D features in complicated interchange scenarios with stacked convolution and the result achieved over 93\% classification accuracy.  
In addition, the new low-cost semi-solid-state LiDAR sensor Livox Mid-40 featuring a incommensurable rosette scanning pattern has demonstrated its potential in high-definition urban mapping.
\end{abstract}


%
\IEEEpeerreviewmaketitle

\section{Introduction}
%
%
%
%
\IEEEPARstart{R}{apid} urbanization and population growth have brought demands and pressure for urban transportation infrastructures. 
Multi-layer interchanges have transformed road intersections from 2D to 3D spaces to reduce traffic flow interference in different directions \cite{chengThreeDimensionalReconstructionLarge2015}.
As a result, multi-layer interchanges have improved traffic efficiency and driving safety. 
Accurate 3D mapping of road infrastructure provides spatial information for various applications, including navigation, traffic management, autonomous driving and urban landscaping \cite{wangLiDARPointClouds2018}.
Multi-layer interchanges are complicated road objects with different designs in various terrain environments, making them difficult to be mapped and modelled. 

Mobile laser scanning (MLS) systems are known to be a useful tool for 3D urban road infrastructure mapping with the capability of collecting high-density and high-accuracy 3D point clouds on moving platforms \cite{maMobileLaserScanned2018}.
Traditional airborne and vehicle-mounted MLS systems usually have difficulties capturing the complicated structure of interchanges due to limitations of flying altitudes or road surfaces \cite{oudeelberink3DInformationExtraction2009}. 
Unmanned aerial vehicle (UAV) is a thriving platform for various sensors, including Light Detection and Ranging (LiDAR) systems, with the flexibility of data collection.
However, there are more uncertainties in point clouds collected by UAV systems, such as point registration and calibration, due to less stable moving positions than aeroplanes and cars.
To capture features from the massive amounts of unordered and unstructured point clouds, deep learning methods have outperformed conventional threshold-based methods and methods using hand-crafted features in recent years  \cite{liDeepLearningLiDAR2020}.
Even though a number of deep learning methods has been developed in recent years, diversified point cloud datasets are still in demand for the semantic segmentation task \cite{gaoAreWeHungry2020}.

This study examines UAV point clouds collected by a new LiDAR sensor on multi-layer stack interchange mapping. 
An improved point-based deep learning architecture was tested to classify the point clouds.
Stacked convolution blocks were implemented to increase the receptive field of convolutions at each layer to deal with the challenges in complicated outdoor point clouds in the stack interchange scene.

\section{Related works}
LiDAR sensors have been widely used for 3D road infrastructure mapping due to its capability of acquiring 3D information, and MLS systems have achieved great success in road inventory extraction \cite{guanUseMobileLiDAR2016}.
However, a large portion of existing methods on road inventory mapping with MLS systems require road surface extraction as the first step \cite{maMobileLaserScanned2018}.
Road surface extraction methods, such as upward-growing algorithms \cite{yuAutomatedExtractionUrban2015} and curb-based algorithms \cite{guanUsingMobileLaser2014}, take advantage of the high-density road points acquired by vehicle-mounted MLS systems.
Airborne Laser Scanning (ALS) systems are the most common sensors for urban stack interchange mapping and reconstruction.
In the early days, due to the limited point density of ALS point clouds, interchange bridge reconstruction relied on topographic maps to determine bridge surface sections \cite{oudeelberink3DInformationExtraction2009}.
Using higher-resolution ALS point clouds with about 1 m spacing, interchange bridge segments can be separately extracted by threshold-based algorithms \cite{chengThreeDimensionalReconstructionLarge2015}.
However, most existing studies on stack interchange mapping only focused on the road surfaces, while the underneath structures like piers and beams were disregarded due to occlusions. 
In this study, the UAV LiDAR point clouds contain flyover bridge structures underneath the bridges, and acquired limited road surface points due to the slant scan angle and heavy traffic.
Therefore, when previous methods focusing on road surfaces may not apply, deep learning methods are used in this study.

Many deep learning frameworks that performs semantic segmentation of LiDAR point clouds could be grouped into the following categories: view-based, voxel-based, point-based and graph-based methods \cite{liDeepLearningLiDAR2020,guoDeepLearning3D2020}. 
View-based methods project point clouds into 2D images to utilize image-based networks \cite{suMultiviewConvolutionalNeural2015,wuSqueezeSegConvolutionalNeural2018}, but occlusions during the projection process would lose valuable 3D features.
Voxel-based methods transform unstructured points into structured voxels to perform 3D convolutions \cite{maturanaVoxNet3DConvolutional2015,huangPointCloudLabeling2016}, but voxels might not be efficient in outdoor scenarios where points are sparse.
Point-based methods directly consume raw point clouds \cite{qiPointNetDeepLearning2017,qiPointNetDeepHierarchical2017,thomasKPConvFlexibleDeformable2019,maMultiScalePointWiseConvolutional2020,huRandLANetEfficientSemantic2020}, but the large number of points pose challenges on the sampling methods and size of receptive field of convolutions to extract features effectively.
Graph-based methods first construct spatial graphs and perform graph convolutions \cite{wangDynamicGraphCNN2019,wangGraphAttentionConvolution2019,liTGNetGeometricGraph2020}, but graph construction could be expensive in computation.
Among the various methods, the point-based network KPFCNN \cite{thomasKPConvFlexibleDeformable2019} has shown superior performance in a number of outdoor point cloud datasets, including Paris-Lille-3D \cite{roynardParisLille3DLargeHighquality2018}, SemanticKITTI \cite{behleySemanticKITTIDatasetSemantic2019} and Toronto-3D \cite{tanToronto3DLargeScaleMobile2020}. 

Outdoor mapping, especially for large road infrastructures, requires the algorithms to capture features at large scales.
Limited by the number of points to consume each time, methods derived from the structure of PointNet \cite{qiPointNetDeepLearning2017} and PointNet++ \cite{qiPointNetDeepHierarchical2017}, including PointSIFT \cite{jiangPointSIFTSIFTlikeNetwork2018} and DGCNN \cite{wangDynamicGraphCNN2019}, do not perform well on outdoor point cloud datasets \cite{maMultiScalePointWiseConvolutional2020,liTGNetGeometricGraph2020}.
To increase receptive field of point feature extraction, multi-scale grouping (MSG) was proposed in  \cite{qiPointNetDeepHierarchical2017}.
However, MSG is generally very time and space-consuming in computation \cite{tanToronto3DLargeScaleMobile2020}.
Stacking point convolution layers can increase the receptive field \cite{engelmannDilatedPointConvolutions2020}, and it has shown promising results in semantic segmentation tasks in outdoor scenarios \cite{huRandLANetEfficientSemantic2020}.
In this study, an improved segmentation network with stacked kernel point convolutions (KPConv) \cite{thomasKPConvFlexibleDeformable2019} is explored in a multi-layer stack interchange scene collected by a UAV LiDAR system.

\section{Dataset}
\begin{figure}[t]
    \centering
    \includegraphics[width=.9\linewidth]{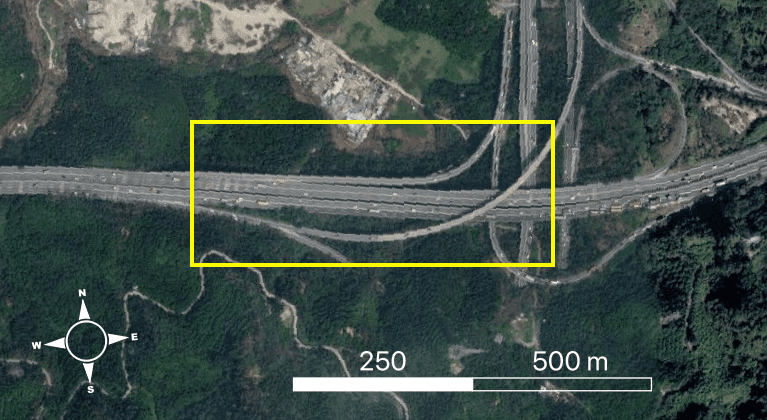}
    \caption{Fulong Flyover and point cloud coverage}
    \label{fig:studyarea}
\end{figure}
The dataset used in this study was a part of Fulong Flyover interchange located in Shenzhen, China. 
Fulong Flyover was built by the northern slope of a mountain, and it connects Henglongshan Tunnel to the south.
A satellite image of Fulong Flyover with the LiDAR point clouds' approximate coverage is illustrated in Fig. \ref{fig:studyarea}. 
The point clouds span approximately 400 m of road segments and cover several flyover bridges.
The UAV flew along the flyover bridge at about 15-20 m distance several times to collect the point clouds.
Livox High-Precision Mapping\footnote{\url{https://github.com/Livox-SDK/livox_high_precision_mapping}\label{livoxgit}} software was used to stitch the scans to produce the point cloud map.

\begin{figure}[t]
    \centering
    \includegraphics[width=.5\linewidth]{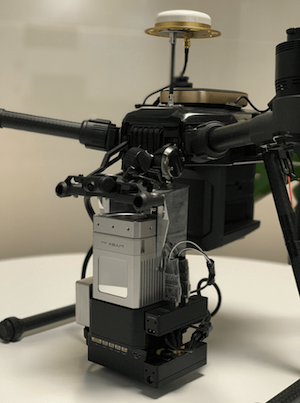}\\
    \caption{UAV LiDAR system with Livox Mid-40\textsuperscript{\ref{livoxgit}}}
    \label{fig:sensor}
\end{figure}

\begin{figure}[t]
    \centering
    \includegraphics[width=.6\linewidth]{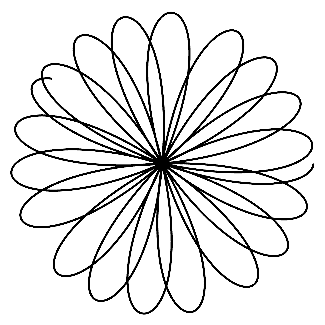}
    \caption{Scanning pattern of Livox LiDAR}
    \label{fig:pattern}
\end{figure}

The MLS system features a Livox Mid-40 LiDAR\footnote{\url{https://www.livoxtech.com/mid-40-and-mid-100}} mounted on an APX-15 UAV\footnote{\url{https://www.applanix.com/products/dg-uavs.htm}}, with a GNSS system, as shown in Fig. \ref{fig:sensor}.
The Livox Mid-40 is a robotic prism-based LiDAR different from conventional multi-line LiDAR commonly used in autonomous vehicles such as Velodyne HDL-32E. 
The new type of low-cost LiDAR sensor features an incommensurable scanning pattern with peak angular density at the center of field-of-view (FOV), which is similar to the fovea in human retina \cite{liuLowcostRetinalikeRobotic2020}.
As illustrated in Fig. \ref{fig:pattern}, the incommensurable rosette scanning pattern will result in denser point clouds over time compared to conventional multi-line LiDARs which scan repeatedly without increased point density. 
Some key technical specifications are listed in \ref{tab:specs}.
Livox Mid-40 has a circular FOV of 38.4\degree with an angular precision over 0.1\degree, and it has a range of up to 260 m with a range precision of 2 cm.
In addition, it is capable of capturing 100,000 points/second. 
Compared with conventional revolving mechanical LiDAR sensors, the Livox LiDAR has a smaller FOV but the incommensuarble scanning pattern could increase the point density over time, filling up to 20\% FOV at 0.1 second and 90\% at 1 second.
Therefore, the Livox LiDAR could acquire points with high density at a fraction of the cost of mechanical LiDARs in some applications where full 360\degree FOV are not required.
For instance, UAV systems do not require data acquisition at 360\degree, and the compact size and light weight of Livox Mid-40 make it suitable for scenarios in this study.
\begin{table}[t]
\caption{Technical specifications of Livox Mid-40}
\label{tab:specs}
\centering
\begin{tabular*}{.9\linewidth}{@{\extracolsep{\fill}}ll}
\toprule \midrule
Wavelength        & 905 nm                    \\ 
Detection range   & 260 m @ 80\% reflectivity \\
Field of view     & 38.4\degree               \\
Range precision   & 2 cm                      \\
Angular precision & 0.1\degree                \\
Acquisition rate  & 100,000 points/s          \\ 
FOV coverage      & 20\% @ 0.1 s, 93\% @ 1 s  \\
Dimensions        & 88 x 69 x 76 mm           \\ \bottomrule   
\end{tabular*}%
\end{table}

\begin{figure*}[t]
    \centering
    \includegraphics[width=.8\linewidth]{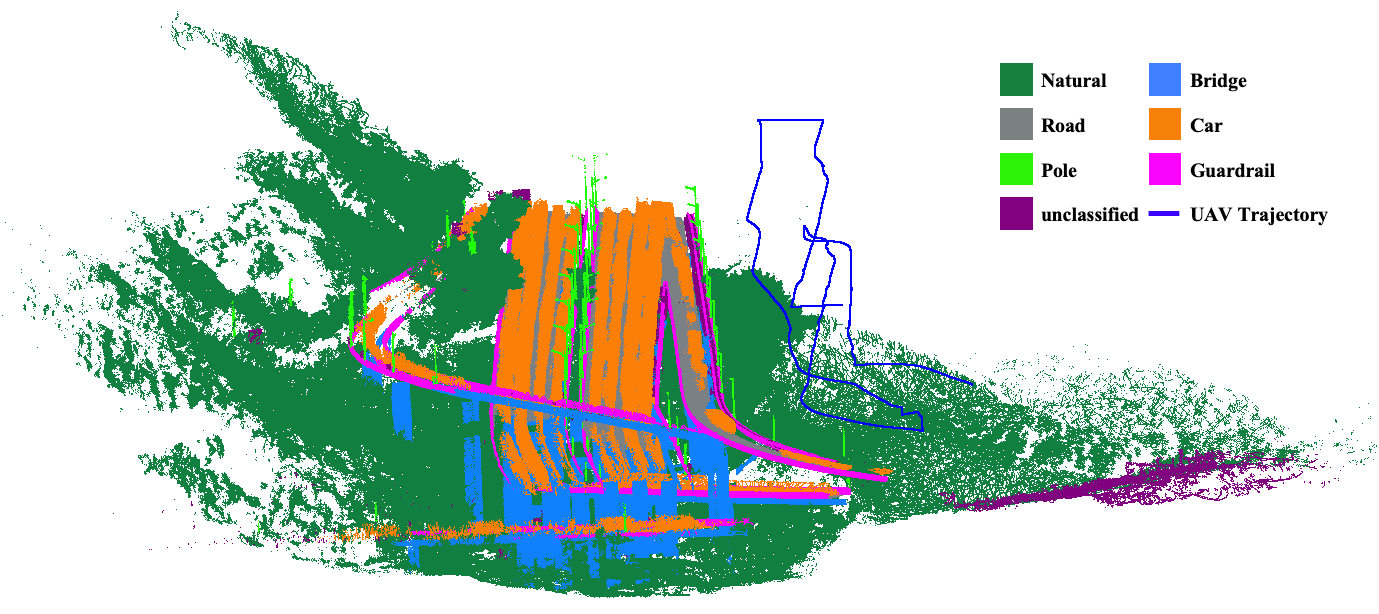}
    \caption{Overview of labelled point clouds and UAV trajectory}
    \label{fig:overview}
\end{figure*}

The dataset consists of over 65 million points with $xyz$ coordinates and intensity.
All the points were manually labelled into 6 classes: natural, bridge, road, car, pole and guardrail.
The classified point cloud and the UAV trajectory are illustrated in Fig. \ref{fig:overview}.
The data was acquired at a slant angle, and traffic was heavy on the flyover bridge, so that little complete road surface segments could be used to evaluate surface density. 
The road surface density at near range could be estimated as 500-1000 points/m\textsuperscript{2}, matching the performance of common vehicle-mounted 32-line LiDAR sensors.

This dataset poses several challenges to point cloud processing algorithms. 
First, the flyover bridges are very close to the mountain, and vegetation appears over, beside and under the bridges. Conventional threshold-based algorithms would be difficult to separate the bridge structure.
Second, the road surface points are rare and incomplete because of the slant angle and heavy traffic at data collection. Algorithms rely on detecting road surfaces may not perform as expected.
Finally, the bridge components are very large objects so that capturing features at a large scale would be challenging.

\section{Proposed method}

The task of semantic segmentation is to assign class labels to each point of the point clouds.
The proposed method is an end-to-end semantic segmentation framework that takes raw point clouds directly and produce point-wise classification labels.

Similar to image convolutions, a convolutional operation at point coordinate $x$ could be generally defined as:
\begin{equation} \label{eq1}
    (\mathcal{F}*g)(x)=\sum_{x_i\in\mathcal{N}_x}g(x_i-x)f_i
\end{equation}
$\mathcal{N}_x$ stands for all the points within a distance of $r$ to the center point $x$.
$x_i$ is a point in space $\mathcal{N}_x$ and $f_i$ is its corresponding features from $\mathcal{F}$. 
Then define \(x'_i=x_i-x\) to be the relative position of $x_i$ to the center point $x$. 
The KPConv $g$ applies different weights $W_k$ to each region with regard to a linear correlation between $x'_i$ and the kernel points $\Tilde{x}_k$:
\begin{equation} \label{eq2}
    g(x'_i)=\sum_{k<K}max(0,1-\frac{||x'_i-\Tilde{x}_k||}{d})W_k
\end{equation}
where $K$ is the total number of kernel points, and $d$ refers to the influence distance.
In this study, same settings as \cite{thomasKPConvFlexibleDeformable2019} were used: \(K=15\), \(d=1.5r\), and the kernel points are located at the tips of a 15-tipped regular polyhedron centered at $x$.

To increase the receptive field of point convolutions in outdoor point clouds, stacked convolution blocks are implemented to capture more point features. 
In the same layer, $n$-time stacked convolutions could capture point features within range $nr$ without the expense of recalculating neighboring points.
There are disagreements on the number of stacked convolutions \cite{huRandLANetEfficientSemantic2020,engelmannDilatedPointConvolutions2020}, and triple KPConv operations were implemented in this study through experiments.

\begin{figure*}[!t]
    \centering
    \includegraphics[width=.9\linewidth]{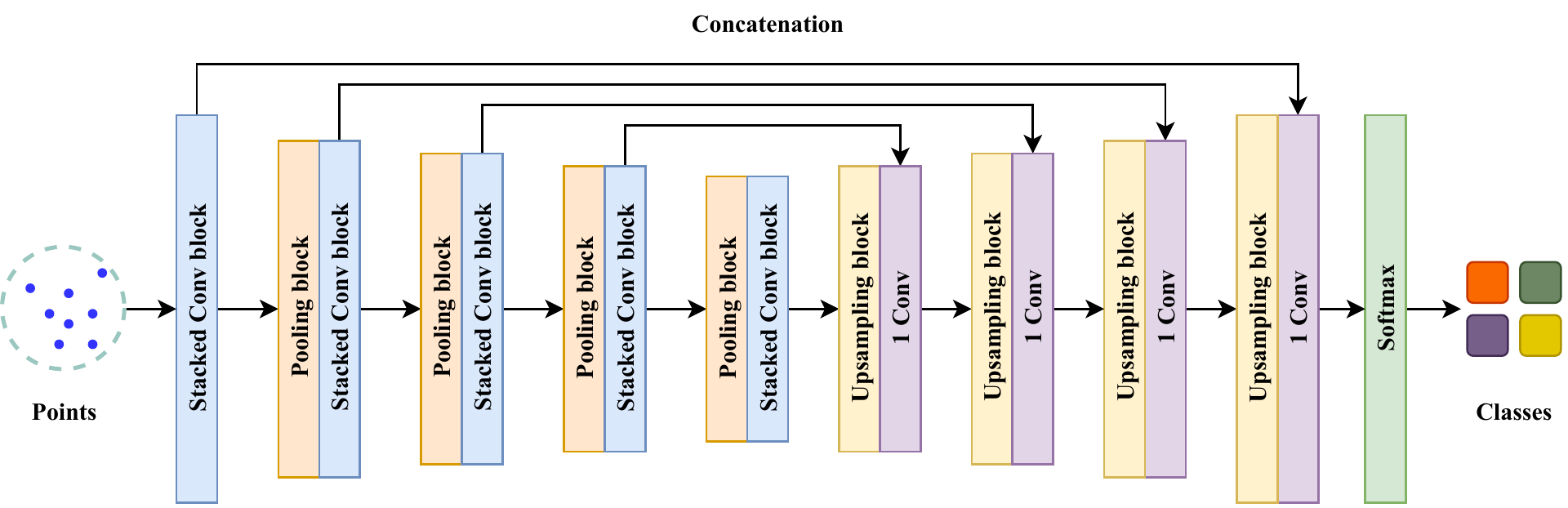}
    \caption{Framework of network architecture for semantic segmentation}
    \label{fig:diagram}
\end{figure*}

The semantic segmentation network architecture is built upon the KPFCNN architecture \cite{thomasKPConvFlexibleDeformable2019}.
As illustrated in Fig. \ref{fig:diagram}, the proposed network uses 5-layer U-Net styled architecture \cite{ronnebergerUNetConvolutionalNetworks2015}.
The architecture mainly contains the following blocks: stacked convolution blocks, pooling blocks and upsampling blocks.
Stacked convolution blocks with three sets of batch normalization-KPConv-Leaky ReLU are implemented to extract point features.
The first layer encodes features at radius $r=0.1$ m, and next four layers containing stacked convolution blocks and pooling blocks captures features at radii \(r=0.2, 0.4, 0.8, 1.6\) m respectively.
The pooling blocks utilize grid subsampling to reduce the number of points, and the features are propagated to the subsampled points via another KPConv operation.
Nearest-neighbor upsampling was used to get the features at the final layers, and skip links pass features at intermediate layers.
Finally, a softmax layer determines the assignment of class of each point.

The more challenging eastern 1/3 section with multiple layers of bridges was selected as the testing set. 
The rest 2/3 of the dataset with fewer layers of bridges and less curved roads was used as the training set. 
The point clouds were resampled into 10 cm grids prior to training, and only point coordinates were used in this study.
Data augmentation methods includes random shuffling, scaling, and random rotation around $z$ axis.
Each step randomly selects one point in the point clouds as the center and collects all points within 5 m range to form a batch.
Cross-entropy loss was used for the segmentation task, and the Momentum Gradient Descend optimizer was used. 
The initial learning rate was set to 0.01, and a momentum of 0.98 was used.
The network was trained on a NVIDIA RTX 2080 Ti with batch size set to 6. 

Intersection over union (\(IoU\)) of each class, overall accuracy (\(OA\)) and mean IoU (\(mIoU\)) are used to evaluate semantic segmentation results.
\begin{equation}
    \label{eq3}
    IoU=\frac{TP}{TP+FP+FN}
\end{equation}
\begin{equation}
    \label{eq4}
    OA=\frac{\sum TP}{N}
\end{equation}
where \(TP\), \(FP\) and \(FN\) represent the numbers of predicted points of true positives, false positives and false negatives respectively, and $N$ stands for the total number of points.
\(IoU\) of each class measures the performance on each class.
\(mIoU\) is the mean of \(IoU\)s of all classes evaluated.
\(OA\) and \(mIoU\) evaluate the overall quality of semantic segmentation.

\section{Results and discussions}
\begin{figure}
    \centering
    \includegraphics[width=\linewidth]{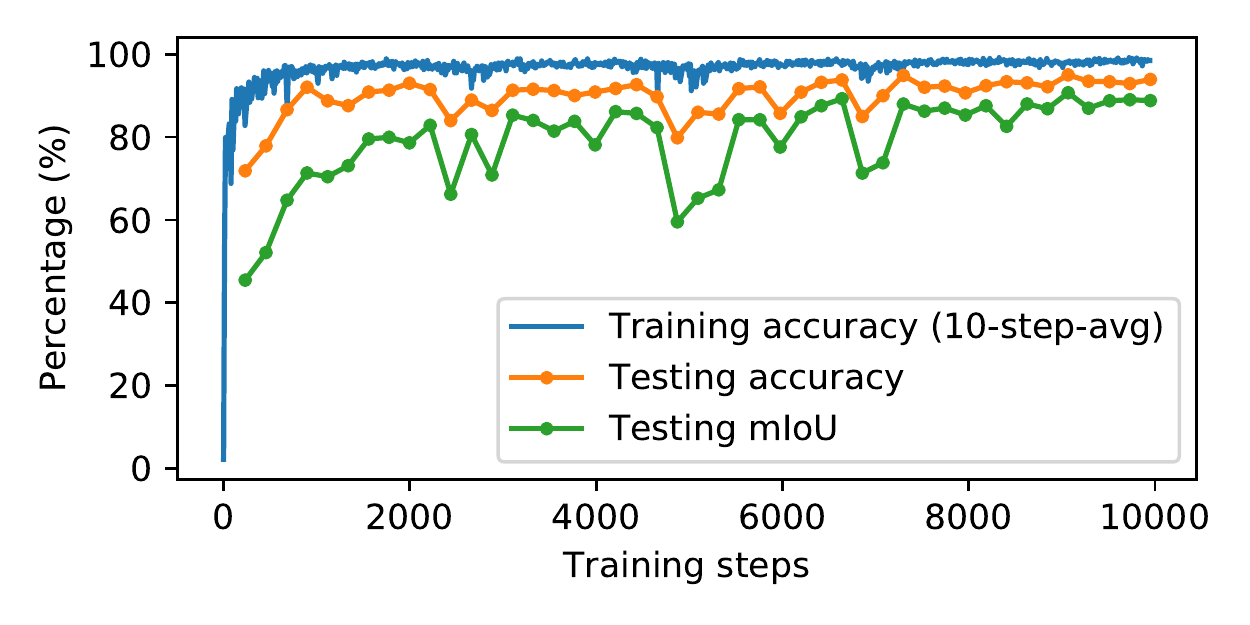}
    \caption{Training and testing curves of the network}
    \label{fig:curve}
\end{figure}

\begin{table*}[t]
\caption{Evaluation of semantic segmentation}
\label{tab:results}
\begin{tabular*}{\textwidth}{@{\extracolsep{\fill}}ccccccccc}
\toprule \midrule
\textbf{Method}             & \textbf{OA}      & \textbf{mIoU}    & \textbf{Natural} & \textbf{Bridge}  & \textbf{Road}    & \textbf{Car}     & \textbf{Pole}    & \textbf{Guardrail} \\ \midrule
PointNet++         & 84.86\% & 70.81\% & 75.98\% & 77.82\% & 83.79\% & 79.70\% & 59.40\% & 48.19\%   \\
MS-TGNet           & 83.80\% & 69.03\% & 78.19\% & 69.43\% & 81.65\% & 65.93\% & 68.24\% & 50.74\%   \\
KPFCNN             & 90.89\% & 82.60\% & 87.25\% & 78.23\% & 88.72\% & 88.55\% & 77.14\% & 75.72\%   \\
Ours - double Conv & 91.59\% & 86.54\% & 86.70\% & 82.71\% & \textbf{90.45\%} & \textbf{95.01\%} & 89.56\% & 74.83\%   \\
\textbf{Ours - triple Conv} & \textbf{93.54\%} & \textbf{88.71\%} & \textbf{90.52\%} & \textbf{86.53\%} & 88.79\% & 93.09\% & \textbf{91.41\%} & \textbf{81.94\% }  \\ \bottomrule
\end{tabular*}
\end{table*}
The dataset is relatively small so that the network converges quickly, and the curves for training accuracy, testing accuracy and testing $mIoU$ are shown in Fig. \ref{fig:curve}.
Table \ref{tab:results} shows the results of the proposed method in comparison to some recent 3D semantic segmentation algorithms.
Compared with KPFCNN \cite{thomasKPConvFlexibleDeformable2019}, our proposed method utilizing stacked convolutions has shown improved results in both $OA$ and $mIoU$.
The most significant improvement has been made on pole identification with 14\% increase in $IoU$.
The performance on car, bridge and road classification also improved by a noticeable margin.
The convolution blocks with triple KPConv operations achieved the highest performance, which agrees with the findings of \cite{engelmannDilatedPointConvolutions2020}.
With an $OA$ of over 93\% and a $mIoU$ over 88\%, the semantic segmentation result of the proposed method could provide confident guidance on high-definition mapping and 3D reconstruction of urban road infrastructures.

\begin{figure}[t]
    \centering
    \includegraphics[width=\linewidth]{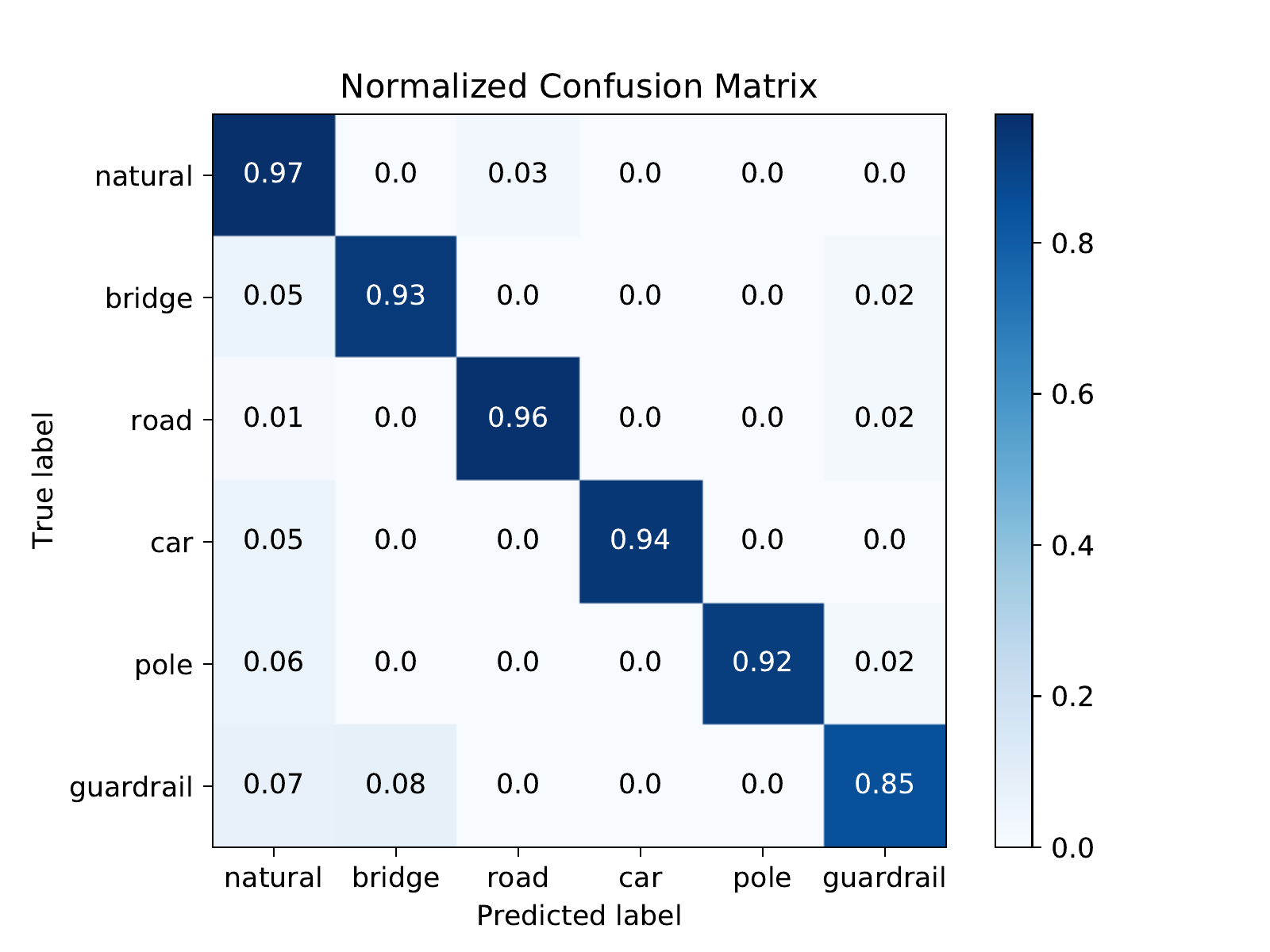}
    \caption{Confusion matrix of semantic segmentation result}
    \label{fig:conf}
\end{figure}

The testing set scene is much more complicated than the training set, with multi-layered bridges with many occlusions. 
The confusion matrix shown in Fig. \ref{fig:conf} provides a detailed look at the sources of error in semantic segmentation.
The significant errors could be attributed to the misclassification of multiple classes to natural. 
The confusion between guardrail and bridge could be the next most significant confusion.


\begin{figure*}[t]
    \centering
    \begin{subfigure}[b]{.45\linewidth}
        \centering
        \includegraphics[width=\linewidth]{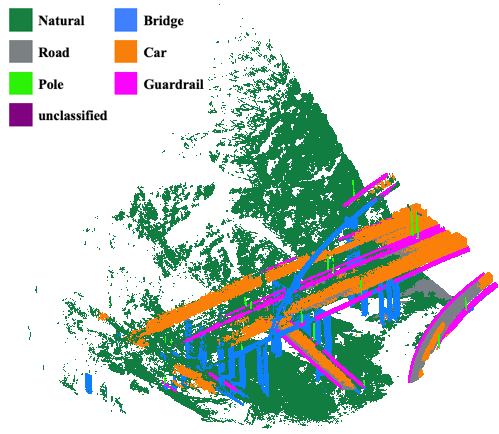}
        \caption{Result of proposed method}
        \label{fig:results}
    \end{subfigure}
    \begin{subfigure}[b]{.45\linewidth}
        \centering
        \includegraphics[width=\linewidth]{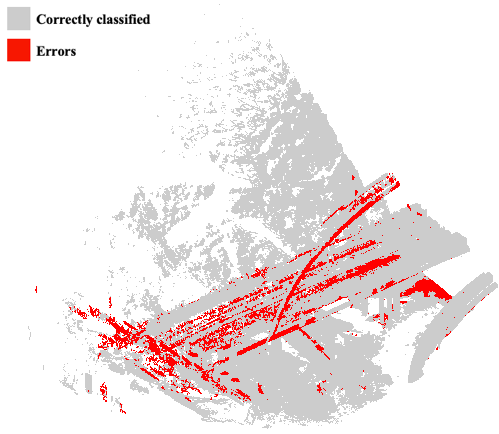}
        \caption{Erroneous points}
        \label{fig:error}
    \end{subfigure} 
        \caption{Visualization of semantic segmentation result}
\end{figure*}

Since the testing scene is in the middle of the stack interchange far from the UAV's trajectory illustrated in Fig. \ref{fig:overview}, the point density at the testing scene is relatively sparse. 
The occlusions by the upper-level bridges make the point density of lower-level bridges even sparser. 
From the comparison of the segmentation result and the erroneous points shown in Fig. \ref{fig:results}, lots of erroneous points could be observed in the left half of the scene at low-level bridge sections.
Besides, the flyover intersection is very close to the mountains and the vegetation, resulting in further difficulties in semantic segmentation. 
In terms of the errors on poles, errors could be found on the two street lamps' lamp parts on the right in Fig. \ref{fig:error}, while the pole parts are correctly classified.
Besides, some flat surfaces underneath the bridge at the right corner were classified as road due to the flatness.

The confusion between guardrail and bridge mostly appeared on the top layer of the interchange. 
The horizontal substructures underneath the bridge surface have a similar structure compared to guardrails in this dataset due to only one side of the bridges was visible to the UAV LiDAR. 
Multi-angle and multi-directional scans from the UAV would potentially increase the performance of semantic segmentation. 

Even though most of the observed errors in this study could be attributed to the underneath structures of the flyover bridges, these structures are often not visible in ALS point clouds acquired from above in previous studies \cite{chengThreeDimensionalReconstructionLarge2015}.
The UAV LiDAR system provides a different view of road infrastructures compared with ALS and vehicle-based systems so that structures underneath the bridges could be mapped and modelled with an additional perspective. 

This study intends to serve as a conceptual and preliminary experiment on stack interchange mapping using the new type of LiDAR sensor.
Some limitations could be addressed in future researches.
First, the dataset is relatively small so that the algorithms are prone to overfit, but this study demonstrated the proposed dataset could achieve accurate semantic segmentation results of the complicated scenarios even with little training data.
Transfer learning could be applied in this small dataset to take advantage of previously trained weights on larger datasets.
Extra data augmentation methods could be applied on the minority classes to improve the robustness of the proposed network.
The proposed algorithm will be more thoroughly evaluated when more data are available.
Second, there are a few calibration issues can be improved to reduce some distortions and artifacts possibly due to vibrations.
LiDAR odometry and mapping (LOAM) algorithms \cite{linLoamLivoxFast2019} could be incorporated in addition to the GNSS information at post processing.
Next, only point coordinates were used, and the LiDAR reflectance could contribute to better semantic segmentation results.
Last but not least, the semantic segmentation can serve as the initial step of scene recognition, and 3D reconstruction methods could be applied to create accurate 3D models and fill data gaps caused by occlusions.

\section{Conclusion}
This study tested a UAV LiDAR system on urban road infrastructure mapping with a case study on semantic segmentation of a multi-layer stack interchange. 
An automated end-to-end point cloud semantic segmentation network utilizing stacked convolutions was proposed on classifying different components of the stack interchange, i.e. natural, bridge, road, car, pole and guardrail.
The method achieved over 93\% overall accuracy and over 88\% $mIoU$, despite the challenges of lack of road surfaces and complicated structures.
The stacked convolutional layers were effective in increasing the receptive fields and improving semantic segmentation performance.
The results could be extended to various tasks, including urban high-definition 3D mapping and 3D model reconstruction.
In addition, this study also showcases the potential capability of the UAV-mounted Livox Mid-40 MLS system on urban high-definition mapping of complicated scenarios.


%



\section*{Acknowledgment}
The authors would like to thank Livox Technology Ltd. for providing the point cloud dataset.

\ifCLASSOPTIONcaptionsoff
  \newpage
\fi



\bibliographystyle{IEEEtran}
%
\bibliography{references}

\begin{thebibliography}{10}
\providecommand{\url}[1]{#1}
\csname url@samestyle\endcsname
\providecommand{\newblock}{\relax}
\providecommand{\bibinfo}[2]{#2}
\providecommand{\BIBentrySTDinterwordspacing}{\spaceskip=0pt\relax}
\providecommand{\BIBentryALTinterwordstretchfactor}{4}
\providecommand{\BIBentryALTinterwordspacing}{\spaceskip=\fontdimen2\font plus
\BIBentryALTinterwordstretchfactor\fontdimen3\font minus
  \fontdimen4\font\relax}
\providecommand{\BIBforeignlanguage}[2]{{%
\expandafter\ifx\csname l@#1\endcsname\relax
\typeout{** WARNING: IEEEtran.bst: No hyphenation pattern has been}%
\typeout{** loaded for the language `#1'. Using the pattern for}%
\typeout{** the default language instead.}%
\else
\language=\csname l@#1\endcsname
\fi
#2}}
\providecommand{\BIBdecl}{\relax}
\BIBdecl

\bibitem{chengThreeDimensionalReconstructionLarge2015}
L.~Cheng, Y.~Wu, Y.~Wang, L.~Zhong, Y.~Chen, and M.~Li,
  ``\BIBforeignlanguage{en}{Three-{{Dimensional Reconstruction}} of {{Large
  Multilayer Interchange Bridge Using Airborne LiDAR Data}}},''
  \emph{\BIBforeignlanguage{en}{IEEE J. Sel. Top. Appl. Earth Obs. Remote
  Sens.}}, vol.~8, no.~2, pp. 691--708, 2015.

\bibitem{wangLiDARPointClouds2018}
R.~Wang, J.~Peethambaran, and D.~Chen, ``\BIBforeignlanguage{en}{{{LiDAR Point
  Clouds}} to 3-{{D Urban Models}}: {{A Review}}},''
  \emph{\BIBforeignlanguage{en}{IEEE J. Sel. Top. Appl. Earth Obs. Remote
  Sens.}}, vol.~11, no.~2, pp. 606--627, 2018.

\bibitem{maMobileLaserScanned2018}
L.~Ma, Y.~Li, J.~Li, C.~Wang, R.~Wang, and M.~Chapman,
  ``\BIBforeignlanguage{en}{Mobile {{Laser Scanned Point}}-{{Clouds}} for
  {{Road Object Detection}} and {{Extraction}}: {{A Review}}},''
  \emph{\BIBforeignlanguage{en}{Remote Sens.}}, vol.~10, no.~10, p. 1531, 2018.

\bibitem{oudeelberink3DInformationExtraction2009}
S.~J. Oude~Elberink and G.~Vosselman, ``\BIBforeignlanguage{en}{{{3D
  Information Extraction}} from {{Laser Point Clouds Covering Complex Road
  Junctions}}},'' \emph{\BIBforeignlanguage{en}{The Photogram. Record}},
  vol.~24, no. 125, pp. 23--36, 2009.

\bibitem{liDeepLearningLiDAR2020}
Y.~Li, L.~Ma, Z.~Zhong, F.~Liu, M.~A. Chapman, D.~Cao, and J.~Li, ``Deep
  {{Learning}} for {{LiDAR Point Clouds}} in {{Autonomous Driving}}: {{A
  Review}},'' \emph{IEEE Trans. Neural Netw. Learning Syst.}, 2020, {DOI}:
  10.1109/TNNLS.2020.3015992.

\bibitem{gaoAreWeHungry2020}
\BIBentryALTinterwordspacing
B.~Gao, Y.~Pan, C.~Li, S.~Geng, and H.~Zhao, ``\BIBforeignlanguage{en}{Are {{We
  Hungry}} for {{3D LiDAR Data}} for {{Semantic Segmentation}}?}''
  \emph{\BIBforeignlanguage{en}{arXiv:2006.04307 [cs]}}, 2020. [Online].
  Available: \url{http://arxiv.org/abs/2006.04307}
\BIBentrySTDinterwordspacing

\bibitem{guanUseMobileLiDAR2016}
H.~Guan, J.~Li, S.~Cao, and Y.~Yu, ``\BIBforeignlanguage{en}{Use of {{Mobile
  LiDAR}} in {{Road Information Inventory}}: {{A Review}}},''
  \emph{\BIBforeignlanguage{en}{Int. J. Image Data Fusion}}, vol.~7, no.~3, pp.
  219--242, 2016.

\bibitem{yuAutomatedExtractionUrban2015}
Y.~Yu, J.~Li, H.~Guan, and C.~Wang, ``\BIBforeignlanguage{en}{Automated
  {{Extraction}} of {{Urban Road Facilities Using Mobile Laser Scanning
  Data}}},'' \emph{\BIBforeignlanguage{en}{IEEE Trans. Intell. Transport.
  Syst.}}, vol.~16, no.~4, pp. 2167--2181, 2015.

\bibitem{guanUsingMobileLaser2014}
H.~Guan, J.~Li, Y.~Yu, C.~Wang, M.~Chapman, and B.~Yang,
  ``\BIBforeignlanguage{en}{Using {{Mobile Laser Scanning Data}} for
  {{Automated Extraction}} of {{Road Markings}}},''
  \emph{\BIBforeignlanguage{en}{ISPRS J. Photogramm. Remote Sens.}}, vol.~87,
  pp. 93--107, 2014.

\bibitem{guoDeepLearning3D2020}
Y.~Guo, H.~Wang, Q.~Hu, H.~Liu, L.~Liu, and M.~Bennamoun, ``Deep {{Learning}}
  for {{3D Point Clouds}}: {{A Survey}},'' \emph{IEEE Trans. Pattern Anal.
  Mach. Intell.}, 2020, {DOI}: 10.1109/TPAMI.2020.3005434.

\bibitem{suMultiviewConvolutionalNeural2015}
H.~Su, S.~Maji, E.~Kalogerakis, and E.~{Learned-Miller}, ``Multi-view
  {{Convolutional Neural Networks}} for {{3D Shape Recognition}},'' in
  \emph{Proc. {IEEE} {ICCV}}.\hskip 1em plus 0.5em minus 0.4em\relax {Santiago,
  Chile}: {IEEE}, Dec. 2015, pp. 945--953.

\bibitem{wuSqueezeSegConvolutionalNeural2018}
B.~Wu, A.~Wan, X.~Yue, and K.~Keutzer, ``{{SqueezeSeg}}: {{Convolutional Neural
  Nets}} with {{Recurrent CRF}} for {{Real}}-{{Time Road}}-{{Object
  Segmentation}} from {{3D LiDAR Point Cloud}},'' in \emph{Proc. {IEEE}
  {ICRA}}.\hskip 1em plus 0.5em minus 0.4em\relax {Brisbane, QLD}: {IEEE}, May
  2018, pp. 1887--1893.

\bibitem{maturanaVoxNet3DConvolutional2015}
D.~Maturana and S.~Scherer, ``{{VoxNet}}: {{A 3D Convolutional Neural Network}}
  for {{Real}}-time {{Object Recognition}},'' in \emph{Proc. {IEEE}
  {IROS}}.\hskip 1em plus 0.5em minus 0.4em\relax {Hamburg, Germany}: {IEEE},
  Sep. 2015, pp. 922--928.

\bibitem{huangPointCloudLabeling2016}
J.~Huang and S.~You, ``Point {{Cloud Labeling Using 3D Convolutional Neural
  Network}},'' in \emph{Proc. {IEEE} {ICPR}}.\hskip 1em plus 0.5em minus
  0.4em\relax {Cancun}: {IEEE}, Dec. 2016, pp. 2670--2675.

\bibitem{qiPointNetDeepLearning2017}
C.~R. Qi, H.~Su, M.~Kaichun, and L.~J. Guibas,
  ``\BIBforeignlanguage{en}{{{PointNet}}: {{Deep Learning}} on {{Point Sets}}
  for {{3D Classification}} and {{Segmentation}}},'' in
  \emph{\BIBforeignlanguage{en}{{IEEE} {CVPR}}}.\hskip 1em plus 0.5em minus
  0.4em\relax {Honolulu, HI}: {IEEE}, Jul. 2017, pp. 77--85.

\bibitem{qiPointNetDeepHierarchical2017}
C.~R. Qi, L.~Yi, H.~Su, and L.~J. Guibas, ``{{PointNet}}++: {{Deep Hierarchical
  Feature Learning}} on {{Point Sets}} in a {{Metric Space}},'' in \emph{Adv.
  NIPS}, 2017, pp. 5099--5108.

\bibitem{thomasKPConvFlexibleDeformable2019}
H.~Thomas, C.~R. Qi, J.-E. Deschaud, B.~Marcotegui, F.~Goulette, and L.~Guibas,
  ``{{KPConv}}: {{Flexible}} and {{Deformable Convolution}} for {{Point
  Clouds}},'' in \emph{Proc. {IEEE} {ICCV}}.\hskip 1em plus 0.5em minus
  0.4em\relax {Seoul, Korea (South)}: {IEEE}, Oct. 2019, pp. 6410--6419.

\bibitem{maMultiScalePointWiseConvolutional2020}
L.~Ma, Y.~Li, J.~Li, W.~Tan, Y.~Yu, and M.~A. Chapman,
  ``\BIBforeignlanguage{en}{Multi-{{Scale Point}}-{{Wise Convolutional Neural
  Networks}} for {{3D Object Segmentation From LiDAR Point Clouds}} in
  {{Large}}-{{Scale Environments}}},'' \emph{\BIBforeignlanguage{en}{IEEE
  Trans. Intell. Transport. Syst.}}, 2020, {DOI}: 10.1109/TITS.2019.2961060.

\bibitem{huRandLANetEfficientSemantic2020}
\BIBentryALTinterwordspacing
Q.~Hu, B.~Yang, L.~Xie, S.~Rosa, Y.~Guo, Z.~Wang, N.~Trigoni, and A.~Markham,
  ``\BIBforeignlanguage{en}{{{RandLA}}-{{Net}}: {{Efficient Semantic
  Segmentation}} of {{Large}}-{{Scale Point Clouds}}},'' in
  \emph{\BIBforeignlanguage{en}{Proc. {IEEE} {CVPR}}}, Jun. 2020, pp.
  11\,108--11\,117. [Online]. Available: \url{http://arxiv.org/abs/1911.11236}
\BIBentrySTDinterwordspacing

\bibitem{wangDynamicGraphCNN2019}
Y.~Wang, Y.~Sun, Z.~Liu, S.~E. Sarma, M.~M. Bronstein, and J.~M. Solomon,
  ``\BIBforeignlanguage{en}{Dynamic {{Graph CNN}} for {{Learning}} on {{Point
  Clouds}}},'' \emph{\BIBforeignlanguage{en}{ACM Trans. Graph.}}, vol.~38,
  no.~5, pp. 1--12, 2019.

\bibitem{wangGraphAttentionConvolution2019}
L.~Wang, Y.~Huang, Y.~Hou, S.~Zhang, and J.~Shan,
  ``\BIBforeignlanguage{en}{Graph {{Attention Convolution}} for {{Point Cloud
  Semantic Segmentation}}},'' in \emph{\BIBforeignlanguage{en}{Proc. {IEEE}
  {CVPR}}}.\hskip 1em plus 0.5em minus 0.4em\relax {Long Beach, CA, USA}:
  {IEEE}, Jun. 2019, pp. 10\,288--10\,297.

\bibitem{liTGNetGeometricGraph2020}
Y.~Li, L.~Ma, Z.~Zhong, D.~Cao, and J.~Li, ``{{TGNet}}: {{Geometric Graph CNN}}
  on 3-{{D Point Cloud Segmentation}},'' \emph{IEEE Trans. Geosci. Remote
  Sens.}, vol.~58, no.~5, pp. 3588--3600, 2020.

\bibitem{roynardParisLille3DLargeHighquality2018}
X.~Roynard, J.-E. Deschaud, and F.~Goulette,
  ``\BIBforeignlanguage{en}{Paris-{{Lille}}-{{3D}}: {{A Large}} and
  {{High}}-quality ground-truth {{Urban Point Cloud Dataset}} for {{Automatic
  Segmentation}} and {{Classification}}},'' \emph{\BIBforeignlanguage{en}{Int.
  J. of Robot. Res.}}, vol.~37, no.~6, pp. 545--557, 2018.

\bibitem{behleySemanticKITTIDatasetSemantic2019}
J.~Behley, M.~Garbade, A.~Milioto, J.~Quenzel, S.~Behnke, C.~Stachniss, and
  J.~Gall, ``{{SemanticKITTI}}: {{A Dataset}} for {{Semantic Scene
  Understanding}} of {{LiDAR Sequences}},'' in \emph{Proc. {IEEE} {ICCV}},
  2019, pp. 9297--9307.

\bibitem{tanToronto3DLargeScaleMobile2020}
W.~Tan, N.~Qin, L.~Ma, Y.~Li, J.~Du, G.~Cai, K.~Yang, and J.~Li,
  ``\BIBforeignlanguage{en}{Toronto-{{3D}}: {{A Large}}-{{Scale Mobile LiDAR
  Dataset}} for {{Semantic Segmentation}} of {{Urban Roadways}}},'' in
  \emph{\BIBforeignlanguage{en}{Proc. {IEEE} {CVPRW}}}, Jun. 2020, pp.
  202--203.

\bibitem{jiangPointSIFTSIFTlikeNetwork2018}
\BIBentryALTinterwordspacing
M.~Jiang, Y.~Wu, T.~Zhao, Z.~Zhao, and C.~Lu,
  ``\BIBforeignlanguage{en}{{{PointSIFT}}: {{A SIFT}}-like {{Network Module}}
  for {{3D Point Cloud Semantic Segmentation}}},''
  \emph{\BIBforeignlanguage{en}{arXiv:1807.00652 [cs]}}, 2018. [Online].
  Available: \url{http://arxiv.org/abs/1807.00652}
\BIBentrySTDinterwordspacing

\bibitem{engelmannDilatedPointConvolutions2020}
\BIBentryALTinterwordspacing
F.~Engelmann, T.~Kontogianni, and B.~Leibe, ``\BIBforeignlanguage{en}{Dilated
  {{Point Convolutions}}: {{On}} the {{Receptive Field Size}} of {{Point
  Convolutions}} on {{3D Point Clouds}}},''
  \emph{\BIBforeignlanguage{en}{arXiv:1907.12046 [cs]}}, 2020. [Online].
  Available: \url{http://arxiv.org/abs/1907.12046}
\BIBentrySTDinterwordspacing

\bibitem{liuLowcostRetinalikeRobotic2020}
\BIBentryALTinterwordspacing
Z.~Liu, F.~Zhang, and X.~Hong, ``Low-cost {{Retina}}-like {{Robotic Lidars
  Based}} on {{Incommensurable Scanning}},'' \emph{arXiv:2006.11034 [cs]},
  2020. [Online]. Available: \url{http://arxiv.org/abs/2006.11034}
\BIBentrySTDinterwordspacing

\bibitem{ronnebergerUNetConvolutionalNetworks2015}
O.~Ronneberger, P.~Fischer, and T.~Brox, ``U-{{Net}}: {{Convolutional
  Networks}} for {{Biomedical Image Segmentation}},'' in \emph{Proc. {MICCAI}},
  N.~Navab, J.~Hornegger, W.~M. Wells, and A.~F. Frangi, Eds.\hskip 1em plus
  0.5em minus 0.4em\relax {Cham}: {Springer International Publishing}, 2015,
  vol. 9351, pp. 234--241.

\bibitem{linLoamLivoxFast2019}
\BIBentryALTinterwordspacing
J.~Lin and F.~Zhang, ``Loam\_livox: {{A}} fast, robust, high-precision
  {{LiDAR}} odometry and mapping package for {{LiDARs}} of small {{FoV}},''
  \emph{arXiv:1909.06700 [cs, eess]}, 2019. [Online]. Available:
  \url{http://arxiv.org/abs/1909.06700}
\BIBentrySTDinterwordspacing

\end{thebibliography}

%








\end{document}